\documentclass[runningheads]{llncs}

\usepackage{eccv}

\usepackage{eccvabbrv}

\usepackage{graphicx}
\usepackage{booktabs}

\usepackage[accsupp]{axessibility}  

\usepackage{hyperref}

\usepackage{orcidlink}
\usepackage{xcolor}
\usepackage{amsmath,amsfonts,bm}
\usepackage{amssymb} 
\usepackage{kotex}
\usepackage{booktabs, adjustbox}
\usepackage[table]{xcolor}
\usepackage{multirow}
\usepackage{booktabs}
\usepackage{adjustbox}
\usepackage{graphicx}

\usepackage{algorithm}
\usepackage{algpseudocode} 
\usepackage{cuted}
\usepackage{capt-of}
\usepackage{pifont}
\newcommand{\cmark}{\ding{51}} 
\newcommand{\xmark}{\ding{55}} 

\newcommand*{\affmark}[1][*]{\textsuperscript{#1}}

\renewcommand{\footnoterule}{%
  \kern -3pt
  \hrule width 0.4\textwidth height 0.4pt
  \kern 2.6pt
}

\begin{document}

\title{Self-Improving Diffusion Classifiers with Minority Preference Optimization} 

\titlerunning{MiPO}

\author{Hyunsoo Kim*\inst{1}\orcidlink{0009-0006-7456-2876} \and Jungmyung Wi*\inst{1}\orcidlink{0009-0005-0913-6566} \and
Soobin Um\inst{2}\orcidlink{0000-0002-1133-0027} \and
Donghyun Kim$^{\dagger}$\inst{1}\orcidlink{0000-0002-7132-4454}\and
Suhyun Kim$^{\dagger}$\inst{3}\orcidlink{0000-0003-0024-1704}}

\authorrunning{H.~Kim et al.} 

\institute{\affmark[1]Korea University, \affmark[2]Kook Min University, \affmark[3]Kyung Hee University \\
    \email{\{climba, wjm333, d\_kim\}@korea.ac.kr} \\
    \email{soobin.um@kookmin.ac.kr}\\
    \email{dr.suhyun\_kim@gmail.com}
}

\maketitle

\begin{NoHyper}
{
\renewcommand{\thefootnote}{\relax}
\footnotetext{$^*$These authors contributed equally to this work.}
\footnotetext{$^\dagger$Co-corresponding author.}
}
\end{NoHyper}

\begin{abstract}
Prior studies have demonstrated that diffusion classifiers achieve robust zero-shot classification performance. However, their effectiveness is strongly tied to the pretraining data distribution: they perform well in majority, high-density regions of the data manifold, but are significantly less accurate in minority, low-density regions. Although prior works on minority sampling have focused on generating more minority-like images, what minority sampling fundamentally enables beyond generation remains underexplored. In this paper, we reveal a direct relationship between minority sampling in generation and the perception capability of diffusion classifiers. Specifically, we show that enhancing minority sampling broadens the coverage of underrepresented regions on the data manifold, thereby improving diffusion-based recognition. To exploit this connection, we propose \textit{Self-Improving Diffusion Classifiers with Minority Preference Optimization} (MiPO), which fine-tunes a pretrained diffusion model using minority preference rewards. Using only arbitrary caption data, MiPO generates candidate samples, rewards those that better cover minority regions, and optimizes the model with LoRA and Group Relative Policy Optimization, without additional image data, external foundation models, or external reward models. This enables stable, prompt-adaptive minority sampling and translates low-density generative coverage into improved zero-shot diffusion classification. To sum up, we show that diffusion classifier perception is biased toward majority regions, demonstrate that this bias can be alleviated through minority preference optimization, and evaluate MiPO on five standard datasets.
\end{abstract}

\section{Introduction}
Beyond the astonishing generative performance of diffusion~\cite{2015diffusion, ho2020denoising, rombach2022high} and flow models~\cite{flowmatching, recflow, sd3}, a new wave of research is emerging that seeks to leverage these generative models to solve various understanding tasks. This trend mirrors the evolution in Natural Language Processing (NLP), where the advent and advancement of large language models (LLMs) resolved numerous downstream tasks such as machine translation~\cite{machinetranslation} or sentiment analysis~\cite{medhat2014sentiment}. Similarly, recent works~\cite{li2023your,wiedemer2025video} leverage large-scale vision generative model to solve vision perception tasks. These studies explore methods such as extracting useful features~\cite{stracke2025cleandift, tang2023emergent}, learned during the generative training process, from the intermediate layers of the models to aid in segmentation or classification, or employing the generative models themselves to solve these tasks directly~\cite{li2023your, wiedemer2025video}.

\label{sec:intro}
\begin{figure}[t]
    \centering
    \includegraphics[width=0.75\columnwidth]{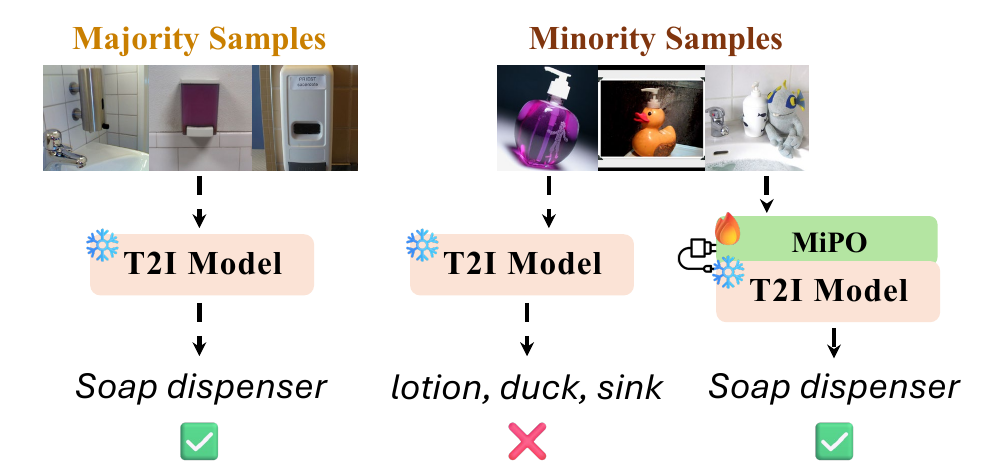}
    \caption{
    \textbf{MiPO expands the perceptual coverage of diffusion classifiers.} A standard diffusion classifier reliably recognizes majority (\ie, high-density) samples that are well covered by the pretrained generative distribution (e.g., \textit{soap dispenser}), but often fails on underrepresented (\ie, low-density) concepts such as \textit{lotion}, \textit{duck}, and \textit{sink}. This illustrates that the recognition ability of diffusion classifiers is closely tied to the generative coverage of the underlying diffusion model. By optimizing minority preference rewards, MiPO encourages the model to better cover such low-density regions, thereby improving the classifier's perception of minority visual categories without requiring additional downstream image data.}

    \label{fig:teaser}
\end{figure}

A prominent example within this direction is the ``Diffusion Classifier''~\cite{li2023your,clark2023dc,chen2025your} which utilizes the generative capabilities of diffusion models to perform zero-shot classification. This approach has garnered significant attention for its distinct advantages, particularly its notable robustness against out-of-distribution (OOD) samples, reduced reliance on spurious shortcut features~\cite{avoidshortcut}, and understanding capability of compositionality~\cite{dccomposition}. However, despite these promising properties, enhancing the performance of generative classifier has not been explored yet.
To tackle this, one fundamental question is: \textbf{``When do the generative models fail at classification?"} Inspired by Richard P. Feynman's assertion, \textit{``What I cannot create, I do not understand,"} our analysis begins by examining the model's generation failures. A widely recognized characteristic of diffusion models is their tendency to excel at generating majority samples from high-density regions, while comparatively struggling with samples from minority (low-density) regions. As illustrated in Fig. ~\ref{fig:teaser}, the performance of the Diffusion Classifier (DC)~\cite{li2023your} on images from minority groups is markedly lower than on those from majority (high-density) groups, measured across various datasets. This phenomenon arises because the noise-matching loss function used to train diffusion models is inherently biased towards these majority modes. While existing research~\cite{um2025minority,um2024self} attempts to address this minority sampling problem by training auxiliary minority classifiers or using prompt optimization to guide the diffusion model, these methods are computationally expensive, requiring additional training or per-prompt optimization. Furthermore, they do not inherently enhance the underlying representation of the diffusion model itself, making it difficult to translate these generation improvements into better diffusion classifier performance.

Therefore, we propose a Self-improving Diffusion Classifier with \textit{\textbf{Minority Preference Optimization}} (MiPO), an effective approach designed to enhance the ability of the diffusion model to represent and generate minority concepts. Prior minority image generation methods~\cite{um2023don,um2024self,um2025minority} primarily focus on generating diverse samples by applying classifier guidance or per-prompt optimization at inference time, without updating the diffusion model itself. As a result, these methods enhance sample diversity but leave the model’s internal representation unchanged, limiting their ability to improve perception tasks such as image classification. In contrast, MiPO enables the diffusion model to self-train its minority representation using only an arbitrary set of prompts, requiring neither images nor external reward models (\eg, CLIPScore~\cite{hessel2021clipscore}, ImageReward~\cite{xu2023imagereward}, PickScore~\cite{kirstain2023pick}, HPS~\cite{wu2023better}) typically used for diffusion alignment. To encourage the generation of minority concepts from these prompts, we leverage online reinforcement learning (\eg, GRPO~\cite{shao2024deepseekmath}) to fine-tune with LoRA by computing group-wise minority reward and updating a reliable generating policy across diffusion timesteps. Furthermore, we employ Kullback–Leibler regularization to enable minority sampling without deviating excessively from the original knowledge of the diffusion model (\ie, preserving majority sampling), allowing the model to generate both majority and minority samples effectively.

Leveraging a LoRA, MiPO offers plug-and-play compatibility, allowing users to attach and detach it from various diffusion backbones (\eg, SD and SD2~\cite{rombach2022high}). Once trained, the adapter enables arbitrary prompt-adaptive minority generation in an efficient manner and allows users to control the strength of minority expression via a LoRA scaling value. 
Beyond improving minority representation, MiPO also brings practical benefits. It adds only a small number of parameters, introduces negligible inference overhead, and requires no prompt-specific optimization, making it suitable for large-scale or real-time generation scenarios. Moreover, because the adapter updates a compact, 
isolated parameter set rather than the full model, it avoids catastrophic shifts in the generative prior and preserves compatibility with downstream alignment or fine-tuning pipelines. This efficient and modular design contributes to the strong performance observed in diffusion-based classification tasks. MiPO demonstrates an 
accuracy gain of over up to 3.8\% on standard benchmarks CIFAR10~\cite{krizhevsky2009learning}, 
ImageNet~\cite{deng2009imagenet})-Tiny as well as on out-of-distribution datasets (CIFAR10-C~\cite{hendrycks2019robustness},  Caltech~\cite{fei2004learning}, SUN09~\cite{choi2010exploiting}), achieving these improvements without using any additional image data and showing particularly substantial gains for minority samples. We believe these are significant findings that can be extended to other classes of generative models. Our contributions can be summarized as follows:

\begin{itemize}
    \item  
    We propose MiPO, a self-improving diffusion classifier that enhances minority representation without using additional training images or external reward models.

    \item   
    MiPO efficiently learns minority preference through GRPO with a compact LoRA adapter, 
    enabling prompt-adaptive minority generation.

    \item  
    MiPO expands the model’s perceptual coverage and achieves significant gains in 
    zero-shot diffusion classification across datasets and backbones.
\end{itemize}

\section{Related Works}
\label{sec:related_works}
\begin{figure*}[t]
    \centering
    \includegraphics[width=\textwidth]{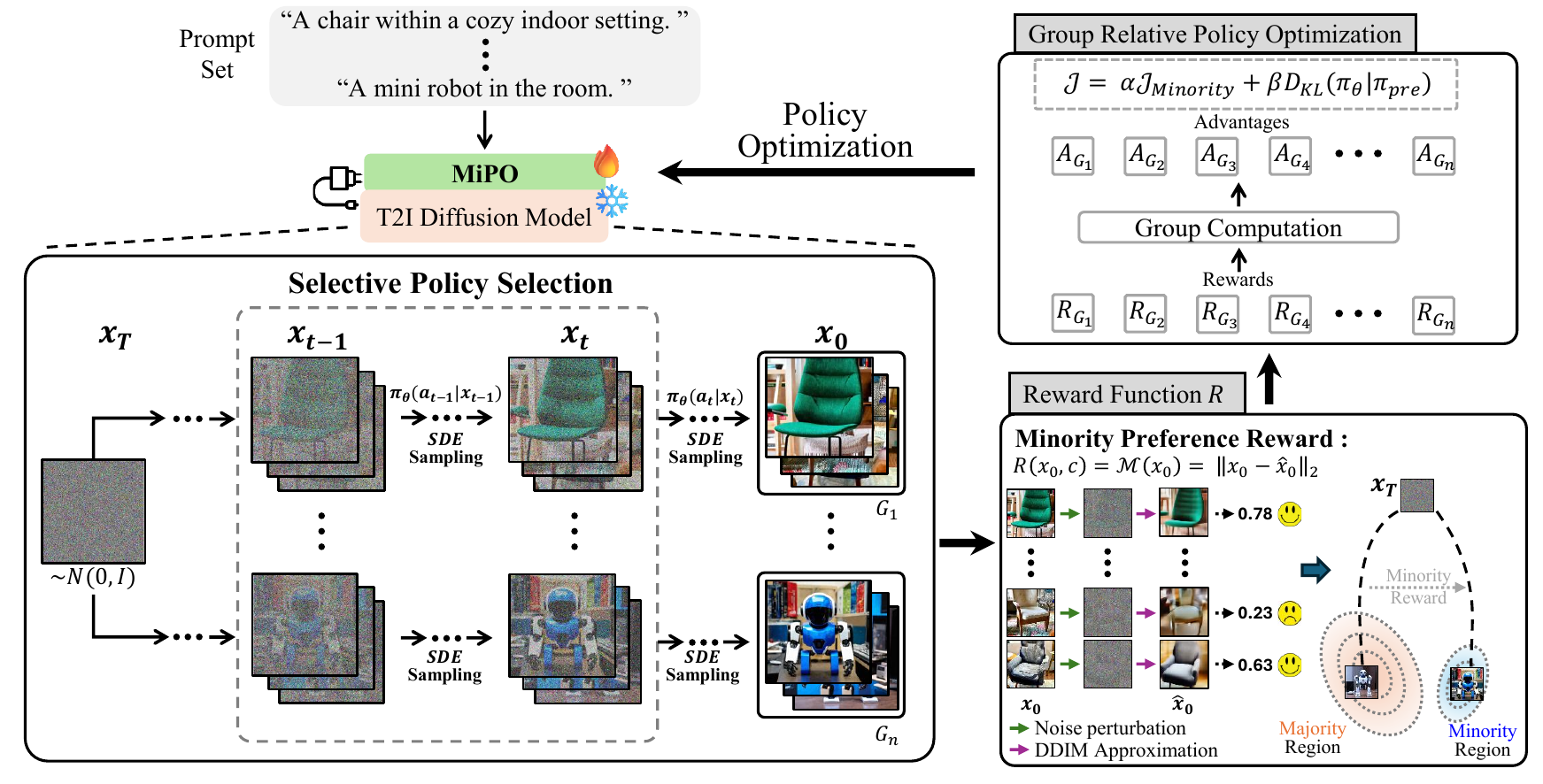}
    \caption{
    \textbf{Training pipeline of Minority Preference Optimization (MiPO).}
    MiPO is integrated into a pretrained text-to-image diffusion model, where multiple SDE trajectories are sampled using the same prompt and initial noise.
    Only the early denoising steps are updated through \textit{Selective Policy Selection}. 
    For each generated sample, the \textit{Minority Preference Reward} is obtained from DDIM-based minority scores, which reflects how strongly the sample lies in a minority region. 
    Rewards within each prompt group are aggregated to compute group-normalized advantages via GRPO, and the final objective combines the minority reward with KL regularization to maintain consistency with the original model. 
    MiPO guides the model toward better coverage of minority regions while preserving a stable denoising.
    }
    \label{fig:pipeline}
\end{figure*}
\noindent\textbf{Minority Sampling with Diffusion Models.} The task of generating minority samples has advanced substantially with the advent of diffusion models, owing to their ability to faithfully capture complex data distributions~\cite{sehwag2022generating, um2023don, um2024self, um2025minority, um2025boostandskip}. Pioneering works in this direction commonly adopt the idea of classifier guidance~\cite{dhariwal2021diffusion}, where separately trained classifiers are employed to steer the generative process toward low-density regions~\cite{sehwag2022generating, um2023don}. Subsequent studies~\cite{um2024self, um2025minority, um2025boostandskip} have addressed practical limitations, such as the dependence on external classifiers~\cite{um2024self}, and have enabled practical minority sample generation that can be performed solely using a pretrained diffusion model across diverse scenarios (\eg, text-to-image generation~\cite{um2025minority}). However, these methods often incur substantial computational overhead during inference due to backpropagation-based guidance for minority representations~\cite{um2023don, um2024self, um2025minority}. While the approach in~\cite{um2025boostandskip} has addressed the complexity issue by developing a guidance-free minority generator, it is limited to stochastic samplers (like DDPM~\cite{ho2020denoising}) and not applicable to deterministic ODE solvers (\eg, DDIM~\cite{song2020denoising}) that are popularly used in modern diffusion-based frameworks. In contrast, our approach is compatible with ODE solvers and requires only marginal additional computation for LoRA construction. Moreover, unlike existing works, we are the first to demonstrate that promoting minority representations can directly enhance the performance of diffusion classifiers, a previously unexplored benefit.

\noindent\textbf{Reinforcement Learning for Generative Model Alignment.}
Recent studies have explored aligning diffusion models with downstream objectives through reinforcement learning based fine-tuning, inspired by the success of RLHF in large language models. 
These methods optimize generated samples according to reward signals while keeping the updated model close to the pretrained distribution. 
Online RL approaches such as DDPO~\cite{black2023training}, DPOK~\cite{fan2023dpok}, PRDP~\cite{deng2024prdp}, AlignProp~\cite{prabhudesai2023aligning}, and DRaFT~\cite{clark2023directly} enable adaptive reward-driven updates but often suffer from instability and high computational cost, whereas offline approaches such as Diffusion-DPO~\cite{wallace2024diffusion}, SPIN-Diffusion~\cite{yuan2024self}, and ID-Aligner~\cite{chen2404id} offer greater stability at the expense of flexibility. 
To mitigate the instability of reward-based optimization, DNO~\cite{tang2024inference} reformulates alignment as a noise-space optimization problem, directly updating denoising trajectories without relying on policy gradients. 
More recently, DanceGRPO~\cite{xue2025dancegrpo} and FlowGRPO~\cite{liu2025flow} further stabilize RL-based diffusion alignment by introducing GRPO~\cite{shao2024deepseekmath} with enabling robust and scalable policy learning across both diffusion and rectified flow models. 
We extend these advances by fine-tuning diffusion models to generate samples from minority regions, enhancing diffusion classifier accuracy.

\noindent\textbf{Diffusion Model for Zero-shot Image Classification.}
Diffusion classifier~\cite{li2023your,clark2023dc} leverages the noise prediction loss of diffusion models to perform zero-shot classification. Given that the noise-matching loss during the diffusion model's training enables the learning of rich image features, DCs not only demonstrate performance comparable to conventional classifiers but are also recognized for distinct advantages, such as robustness to OOD datasets and avoidance of spurious shortcut features~\cite{avoidshortcut, chen2023robust}. Furthermore,~\cite{dccomposition} revealed that DC possess an understanding of compositionality. More recently,~\cite{noisematters} explored the role of noise in DC, investigating the concept of ``good noise." However, these studies do not fundamentally explore methods for improving the intrinsic performance of the diffusion classifier itself. In this paper, we are the first to propose a simple yet effective method for enhancing the performance of the diffusion classifier by strengthening its minority generation capabilities.

\section{Method}

\subsection{Preliminaries}
\label{sec:preliminaries}

\subsubsection{Fine-tuning Diffusion Models with Reinforcement Learning.}
Following recent works~\cite{black2023training,liu2025flow,xue2025dancegrpo}, the denoising trajectory of a diffusion model can be formulated as a Markov Decision Process (MDP), where the policy $\pi_{\theta}(a_{t} \mid x_t) =p_\theta(\bm{x}_{t-1}|\bm{x}_t,\bm{c})$ sequentially generates denoised samples conditioned on the state $s_t = (\bm{c}, t, \bm{x}_t)$. 
For each prompt $\bm{c}$, a group of $G$ trajectories $\{\tau^{(g)} = (\bm{z}^{(g)}_T, \dots, \bm{z}^{(g)}_0)\}_{g=1}^{G}$ is sampled from the same initial noise under stochastic SDE solvers. Each trajectory produces a final output $\bm{z}^{(g)}_0$ and reward $r^{(g)} = R(\bm{z}^{(g)}_0, \bm{c})$.
The group-relative advantage is then computed as
\begin{equation}
A^{(g)} =
\frac{
r^{(g)} - \mathrm{mean}\!\big(\{r^{(j)}\}_{j=1}^{G}\big)
}{
\mathrm{std}\!\big(\{r^{(j)}\}_{j=1}^{G}\big)
},
\label{eq:adv_dancegrpo}
\end{equation}
where the group corresponds to multiple SDE sampling trajectories sharing the same prompt
and initialization noise.
This group-level normalization ensures that policy updates remain invariant to reward scale
and the stochastic variance of different SDE solvers.

The policy is then updated by maximizing the clipped GRPO objective:
\begin{equation}
\begin{aligned}
\mathcal{J}(\theta)
= \mathbb{E}\Bigg[
&\frac{1}{G}\sum_{i=1}^{G}\frac{1}{T}\sum_{t=1}^{T}
\min\!\Big(
\rho_{t,i}(\theta)A_{t}^{(i)},
\\[-2pt]
&\qquad\qquad\qquad
\mathrm{clip}\big(\rho_{t,i}(\theta),\,1-\epsilon,\,1+\epsilon\big)A_{t}^{(i)}
\Big)\Bigg].
\label{eq:grpo_objective}
\end{aligned}
\end{equation}

where $\rho_{t,i}(\theta)=
\tfrac{\pi_\theta(a_{t,i} \mid x_{t,i})}
{\pi_{\theta_{\text{old}}}(a_{t,i} \mid x_{t,i})}$ 
is the likelihood ratio between current and previous policies. 
This group-relative normalization makes the optimization invariant to the absolute scale of rewards and heterogeneous prompt distributions, 
enabling stable and consistent RL-based fine-tuning of diffusion models.

\subsubsection{Diffusion Classifier.}
Diffusion classifier~\cite{li2023your,clark2023dc} is a framework that applies pre-trained diffusion models to classification tasks. Since calculating the class-conditional likelihood $p(x|y)$ in diffusion models is intractable~\cite{li2023your}, DC uses the noise prediction error as a surrogate objective. This is based on the intuition that the noise prediction error will be smaller when conditioned on the correct class.The classification process is as follows: A noisy image $x_t$ is generated from an original image $x_0$ and random noise $\epsilon$. The model then predicts the conditioned noise $\hat{\epsilon}_\theta(x_t, c_i, t)$ for each possible class $c_i$. The final predicted class $\hat{c}$ is the one that minimizes the L2 distance between the predicted noise and the ground-truth noise $\epsilon$.$$\hat{c} = \underset{c_i}{\arg\min} \left\| \hat{\epsilon}_\theta(x_t, c_i, t) - \epsilon \right\|_2^2$$

However, diffusion models are known to favor majority regions because the standard noise-matching objective emphasizes reconstruction of frequently occurring samples~\cite{sehwag2022generating,um2023don,um2024self,um2025minority}.
As a result, rare or underrepresented concepts, referred to as \textit{minority regions}, are often neglected, leading to biased generation and degraded performance of the diffusion classifiers~\cite{li2023your,clark2023dc}.

\subsection{Minority Preference Optimization (MiPO) for Diffusion Classifier}

Building on the formulation in Sec.~\ref{sec:preliminaries}, we now describe how diffusion models can be fine-tuned to better capture underrepresented, minority regions of the data distribution. The previous works~\cite{sehwag2022generating,um2023don,um2024self,um2025minority} introduce inference-time guidance methods that focus solely on generating diverse or minority images without updating the model itself.
In contrast, we aim to go beyond merely generating diverse samples, instead enhancing perceptual capability of diffusion model by strengthening its intrinsic understanding of minority regions. 
To achieve this, we fine-tune the diffusion model with GRPO~\cite{shao2024deepseekmath,deepseekai2025deepseekr1incentivizingreasoningcapability}, 
\textbf{and, unlike the prior works, we apply GRPO updates through a LoRA~\cite{hu2022lora}, enabling efficient and stable policy learning without modifying the full diffusion backbone.}
We adopt a minority preference reward design that encourages minority sampling while preserving the stability and fidelity of the pretrained model. 
The overall training process is illustrated in Fig.~\ref{fig:pipeline}, which depicts the selective policy updates, group-relative optimization, and KL regularization used in MiPO. MiPO does not require any additional real data; instead, it improves generalization by self-generating training samples and refining itself in a self-improving method with minority preference rewards.

\subsubsection{Minority Preference Reward.}
Inspired by recent inference-time guidance methods for minority generation~\cite{um2023don,um2024self,um2025minority},
we adapt their intuition into a training-time optimization framework.
To encourage the diffusion model to generate samples from minority regions of the data distribution,
we introduce a \textit{minority preference reward} that quantifies how underrepresented a generated sample is. 
Given a generated image $\bm{x}_0$ conditioned on prompt $\bm{c}$, 
the reward is computed as 
$R(\bm{x}_0,\bm{c})=\mathcal{M}(\bm{x}_0)$, 
where $\mathcal{M}(\cdot)$ measures the degree of minority.

We compute $\mathcal{M}(\bm{x}_0)$ using a DDIM-based approximation of the Tweedie’s formula~\cite{chung2022diffusion,kotz2012breakthroughs}. 
Specifically, for a generated image $\bm{x}_0$, we add Gaussian noise to obtain a perturbed latent
$\bm{x}_t = \sqrt{\bar{\alpha}_t}\bm{x}_0 + \sqrt{1-\bar{\alpha}_t}\bm{\epsilon}$,
where $\bm{\epsilon}\!\sim\!\mathcal{N}(\mathbf{0},\mathbf{I})$.
Following the common practice in a diffusion study~\cite{um2023don}, 
we set the perturbation strength to an intermediate diffusion step (\eg, $t = 0.9T$) to ensure that the denoiser uncertainty is sufficiently exposed while preserving perceptual structure. 
We then reconstruct $\hat{\bm{x}}_0$ via the DDIM~\cite{song2020denoising}:
\begin{equation}
\hat{\bm{x}}_0 =
\frac{\bm{x}_t - \sqrt{1-\bar{\alpha}_t}\,\bm{\epsilon}_{pre}(\bm{x}_t,c,t)}
{\sqrt{\bar{\alpha}_t}},
\label{eq:ddim_inverse}
\end{equation}
The minority score is then defined as the reconstruction discrepancy between the generated sample 
and its DDIM-approximated posterior mean:
\begin{equation}
\mathcal{M}(\bm{x}_0)=\|\hat{\bm{x}}_0-\bm{x}_0\|_2.
\label{eq:minority_score}
\end{equation}
Intuitively, samples with higher reconstruction errors indicate regions where the denoiser exhibits greater uncertainty or weaker representation, 
corresponding to minority areas of the data manifold. 
In practice, $\mathcal{M}(\bm{x}_0)$ can also be combined with feature-space density estimation, 
LPIPS~\cite{zhang2018unreasonable} based classifier-based likelihood proxies which further refine minority sensitivity by leveraging complementary perspectives of data rarity.  
This reward formulation encourages the model to explore and synthesize samples that lie in sparsely populated regions of the training distribution, leading to enhanced coverage of long-tail semantics and improved diffusion-classifier accuracy.

To integrate this objective into GRPO training, 
the group-normalized advantage in Eq.~\ref{eq:adv_dancegrpo} is computed using the minority reward $R(\bm{x}_0,\bm{c})=\mathcal{M}(\bm{x}_0)$. 
Hence, higher-reward (\ie, minority) samples yield positive advantages, 
while low-reward (\ie, majority) samples produce negative ones, 
guiding the policy toward minority-aligned generation behavior. 
This design ensures that the policy optimization remains prompt-adaptive 
and that minority exploration emerges organically from the reward signal alone.

\subsubsection{Robust Fine-tuning with KL Regularization.}
While the minority reward encourages exploration in minority regions, 
it can also induce instability or degrade visual fidelity if the policy diverges too far from the pretrained distribution. 
To mitigate this issue, we incorporate a Kullback–Leibler (KL) regularization term during GRPO updates:
\begin{equation}
    \mathcal{J}_{\text{}}(\theta)
    = \alpha\mathcal{J}_{Minority}(\theta)
    + \beta D_{\mathrm{KL}}\!\big(\pi_\theta \,\|\, \pi_{\mathrm{pre}}\big),
    \label{eq:ours_objective}
\end{equation}
where $\pi_{\mathrm{pre}}$ denotes the pretrained diffusion model 
and $\beta$ controls the regularization strength. 
Here, $\alpha$ determines the contribution of the minority reward, while $\beta$ regulates how closely the updated policy should remain aligned with the pretrained prior. 
We set $\alpha=0.7$ and $\beta=0.15$ to balance minority enhancement and fidelity preservation. 
This term penalizes excessive deviations from the reward optimization, acting as the constraint that preserves the generative prior of the base diffusion model. 
Empirically, KL regularization stabilizes reward-driven optimization and prevents reward hacking while enabling reliable minority sampling. 
Ablation results on the KL term are presented in Sec.~\ref{sec:grpo_ablation}.
\subsection{Selective Policy Optimization}

Although reinforcement learning can, in principle, update the policy across all denoising timesteps, recent works suggest that optimizing over a subset of timesteps is sufficient for effective diffusion fine-tuning. DanceGRPO~\cite{xue2025dancegrpo} demonstrated that stable policy learning can be achieved even when updates are applied to only a fraction of the diffusion steps, showing that the policy does not need to be optimized across the entire trajectory to obtain performance gains. 
Similarly, ~\cite{chan2025mgd} introduced a \textit{stop guidance} mechanism, highlighting that early-stage timesteps play a dominant role in determining sample quality and that applying guidance at later stages provides diminishing returns.  

Motivated by these findings, we update the policy only for the initial $40\%$ of diffusion timesteps. This selective optimization not only reduces computational cost but also focuses learning on the early denoising phase, 
where coarse semantic and structural information is formed.  Empirically, we find that restricting updates to these earlier timesteps preserves visual fidelity while maintaining strong reward optimization performance, demonstrating that fine-tuning need not encompass the entire diffusion trajectory to yield effective alignment. Ablation results for our selective policy optimization strategy are presented in Sec.~\ref{sec:grpo_ablation}.


\section{Experiments}
We evaluate our approach through three complementary perspectives: 
(i) zero-shot diffusion classification across multiple datasets, 
(ii) ablation studies that analyze the effect of KL regularization and timestep selection, 
and (iii) minority image generation to assess whether the learned policy captures minority visual concepts. 
This structure provides a comprehensive understanding of how minority policy optimization improves both the perceptual ability of the diffusion model and its capacity to represent minority samples.

\subsection{Experimental Setup}
\subsubsection{Training Datasets.}
We employed captions from HPSv2~\cite{wu2023human} for minority policy optimization without any images. 
The HPSv2 caption set is derived from DiffusionDB~\cite{wang2023diffusiondb} and MSCOCO~\cite{lin2014microsoft}, consisting of over 100k cleaned and de-biased text prompts that were curated for large-scale human preference evaluation.
It is noteworthy that our approach requires only arbitrary caption data without needing additional images. 
\begin{table*}[t]
    \centering
    \caption{
    \textbf{Zero-shot classification accuracy of diffusion classifiers on SD~1.5 and SD~2.0.}
    MiPO improves performance across most datasets for both SD~1.5 and SD~2.0, notably achieving gains of about 2.0\% on CIFAR10-C and 1.2\% on ImageNet-Tiny. While CIFAR10 under SD~2.0 shows a slight decrease, the overall trend demonstrates that minority-aware fine-tuning generally enhances robustness and extends perceptual coverage of the diffusion models across diverse domains.
    }
    \normalsize
    \begin{tabular}{l|cccccc}
    \toprule
      Model  & CIFAR10 & CIFAR10-C & ImageNet-Tiny & Caltech & SUN09 \\
    \midrule
    SD1.5         &  85.38 & 57.72  & 46.50 &  97.81  & 64.69    \\
    \rowcolor{gray!10}
    MiPO           &  \textbf{87.89} & \textbf{59.76}  &  \textbf{47.30} &  \textbf{99.51}  &  \textbf{68.25}    \\
    \midrule
    SD2.0           & \textbf{88.58}  & 64.61  & 51.45 & 99.65     & 64.72    \\
    \rowcolor{gray!10}
    MiPO          & 86.98  & \textbf{65.33}  & \textbf{55.25}  & \textbf{99.93} & \textbf{69.26}     \\
    \bottomrule
    \end{tabular}
    \label{tab:sd_version}
\end{table*}

\subsubsection{Evaluation Datasets.}
We utilized the CIFAR10~\cite{krizhevsky2009learning} and Tiny-ImageNet~\cite{le2015tiny} validation sets to measure the accuracy of the diffusion classifier. Furthermore, we evaluated our method on robustness benchmarks, specifically CIFAR10-C~\cite{hendrycks2019robustness}, and Caltech~\cite{fei2004learning}, LabelMe~\cite{russell2008labelme}, SUN09~\cite{choi2010exploiting}, and PASCAL VOC~\cite{everingham2010pascal} in VLCS~\cite{fang2013unbiased} to assess performance against corruptions and perturbations. Due to the significant computational overhead of diffusion classifier evaluation on CIFAR10-C and Tiny-Imagenet, we perform measurements on a random subset of 2,000 images. For the generation and evaluation of minority images, we randomly sampled 1,000 captions from the MS-COCO~\cite{lin2014microsoft} validation set. 
\begin{table*}[t]
    \centering
    \caption{\textbf{Effect of the KL regularization.} 
    Introducing KL regularization leads to consistently better performance 
    for both minority and majority groups, indicating that constraining the 
    policy to stay close to the pretrained diffusion prior stabilizes MiPO.}
    \small
    \begin{tabular}{l l c c}
    \toprule
    \multicolumn{2}{c}{} & \textbf{Minority group} & \textbf{Majority group}  \\
    \midrule
    & SD1.5                  & 73.07   & 79.18  \\
    \midrule
    & w/o KL                 & 78.58   & 82.08   \\
    \rowcolor{gray!10}
    & \textbf{w/ KL (Ours)}  & \textbf{80.58} & \textbf{82.68} \\
    \bottomrule
    \end{tabular}

    \label{tab:ablation-acc-only}
\end{table*}
In addition, to intuitively demonstrate how the diffusion model understands majority and minority samples, we further divide the CIFAR10 dataset into two groups of 1,000 images each, with one representing the majority (\ie, majority group) and the other representing the minority (\ie, minority group), and evaluate the classifier performance separately on both groups. A detailed description of the evaluation dataset and additional dataset configurations is provided in the Appendix.

\subsubsection{Baselines and Evaluation Metrics.}
To the best of our knowledge, this work is the first to improve the diffusion classifier in a self-improving method without using any additional image data such as few-shot examples. 
Hence, we adopt the original diffusion classifier as the main baseline. 
Ablation studies on the proposed components (\eg, minority reward, KL regularization, and selective policy optimization) are presented in Sec.~\ref{sec:futher}. 
Further analyses on the reinforcement learning design choices and additional ablations are provided in the Appendix. 
For minority image generation, we compare our method with two baselines: vanilla Stable Diffusion and Minority Prompt, which represents the current state of the art in minority sampling.

For evaluation metrics, we follow the standard diffusion classifier protocol and report zero-shot classification accuracy for each dataset. 
For minority image generation, we assess the perceptual and preference alignment of generated images using CLIPScore~\cite{hessel2021clipscore}, ImageReward~\cite{xu2023imagereward}, HPSv2~\cite{wu2023human}, and HPSv3~\cite{ma2025hpsv3}. 
To quantify the degree of minority representation in the generated images, we additionally measure Minority Score~\cite{um2023don} and JEPA-SCORE~\cite{balestriero2025gaussian}.

\subsection{Zero-shot Classification Results}
Our proposed method demonstrates consistent performance improvements in diffusion classifier accuracy across all primary evaluation datasets. 
Notably, these gains are achieved without using any additional image data, relying solely on arbitrary captions (HPSv2) and the minority preference as a reward signal. 
As summarized in Table~\ref{tab:sd_version}, MiPO consistently improves zero-shot classification accuracy across both SD~1.5 and SD~2.0. 
For SD~1.5, all datasets exhibit clear improvements, while for SD~2.0 we observe gains of about 2\% on CIFAR10-C and roughly 1.2\% on ImageNet-Tiny, with Caltech and SUN09 showing similar upward trends. 
Although CIFAR10 under SD~2.0 shows a slight decrease, the overall pattern still indicates that MiPO enhances the perceptual coverage of diffusion models and generalizes effectively across diverse domains.

Interestingly, we also observe performance degradation on datasets such as LabelME and VOC2007 (Table~\ref{tab:sd_version_labelme_voc}). These datasets contain noisy, multi-object, or inconsistently annotated images, making them substantially harder and inherently unstable for zero-shot diffusion classification. Since they are not strictly clean benchmarks, we provide a detailed analysis and dataset-specific denoising procedures in the Appendix.

\begin{table}[t]
  \centering
    \caption{\textbf{Ablation study on time step selection.} 
    We compare the time step selection strategies, Full (0--50), Late (30--50), Middle (15--35), and Early (0--20, ours).  Updating the early timesteps results in the most effective policy optimization and achieves the best overall performance.}
  \normalsize
  \resizebox{0.6\linewidth}{!}{
    \begin{tabular}{l l c c}
      \toprule
      \multicolumn{2}{c}{} & \textbf{Minority group}  & \textbf{Majority group} \\
      \midrule
      & SD1.5                  & 73.07   & 79.18  \\
      \midrule
      & Full \,(0--50) & \underline{80.58} & \textbf{82.68} \\
      & Late \,(30--50)              & 73.77 & 78.18 \\
      & Middle \,(15--35)                & 79.90 & 78.28 \\
      \rowcolor{gray!10}
      & \textbf{Early \,(0--20, Ours)}                  & \textbf{81.68} & \underline{82.48} \\
      \bottomrule
    \end{tabular}
  }

  \label{tab:timestep-ablation}
\end{table}

\begin{figure*}[t]
    \centering
    \includegraphics[width=1\linewidth]{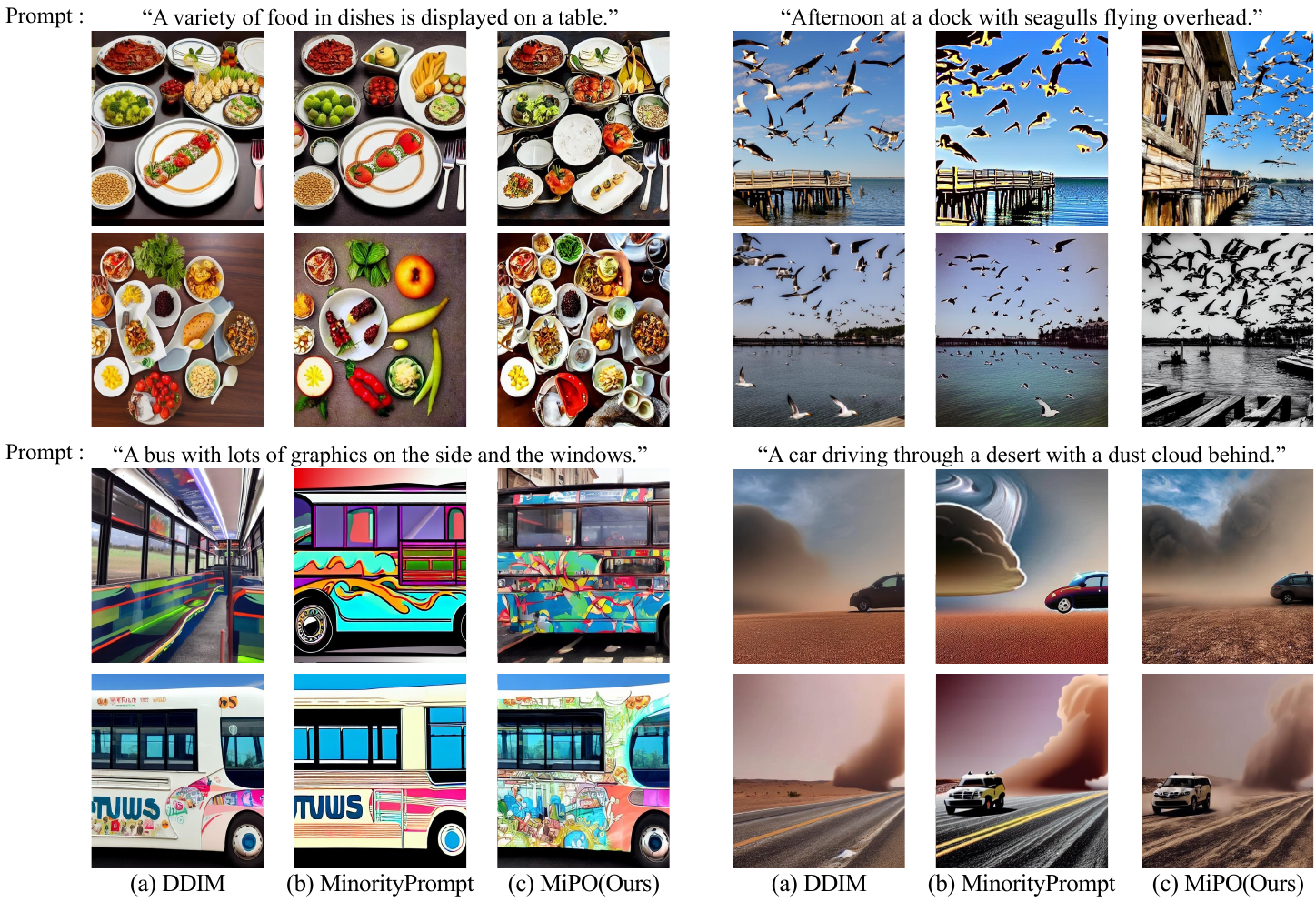}
    \caption{
    \textbf{Qualitative results of minority samples.}
    Across diverse MSCOCO validation prompts, MiPO produces images that exhibit richer minority details and greater concept diversity compared to DDIM, while remaining competitive with  the prompt-optimized MinorityPrompt baseline. Unlike MinorityPrompt, MiPO achieves these results in a fully prompt-adaptive method without per-prompt optimization. \textbf{Images are not cherry-picked.}
    }
    \label{fig:qualitative_results}
\end{figure*}

\begin{table*}[t]
\centering
\caption{
\textbf{Quantitative comparison of minority-oriented image generation on SD~1.5.}
\textit{Minority Gen} indicates whether a method can generate minority samples; 
\textit{DC Eval} denotes compatibility with diffusion-classifier evaluation; 
\textit{Prompt-adaptive} reflects whether the method works without per-prompt optimization at inference. 
Across all quantitative metrics, our MiPO approach achieves the second-best performance while running at the same speed as DDIM (1s per image), unlike MinorityPrompt, which requires heavy prompt-specific optimization (63s). 
These results show that MiPO enables efficient minority generation while maintaining strong perceptual alignment.
}
\resizebox{\linewidth}{!}{
\begin{tabular}{lccc|cccc|c}
\toprule
Method & Minority Gen. & DC Eval. & Prompt-adaptive & CLIPScore $\uparrow$ & HPSv3 $\uparrow$ & Minority Score $\uparrow$ & JEPA-SCORE $\downarrow$ & Time(s) $\downarrow$\\ 
\midrule
DDIM~\cite{song2020denoising}               & \xmark & \cmark & \cmark & \textbf{31.7571}  & \textbf{5.7581} & 0.2132 & -568.7979 &  \textbf{1} \\
MinorityPrompt~\cite{um2025minority}        & \cmark & \xmark & \xmark & 30.4351 & 1.5324 & \textbf{0.2311} & \textbf{-647.9724} &  63  \\ 
\rowcolor{gray!10}
Ours (DDIM + MiPO)                          & \cmark & \cmark & \cmark & \underline{30.8026} & \underline{4.4090} & \underline{0.2289} & \underline{-646.0296} & \textbf{1} \\
\bottomrule
\end{tabular}
}

\label{tab:main_results}
\end{table*}

\section{Further Analyses}
\label{sec:futher}
\subsection{Ablation Study}
\label{sec:grpo_ablation}
To verify whether enhancing minority representation genuinely improves diffusion classifier performance,
we conduct ablation studies using the CIFAR-10 dataset by dividing it into minority and majority groups.
For a fair comparison, we define these groups using a k-Nearest Neighbors (KNN)--based density estimator rather
than the minority score used during training.
This setup ensures that improvements are not an artifact of our reward function but reflect real gains in
representation quality.
Furthermore, prior minority sampling approaches~\cite{um2023don, um2025minority}
have not been shown to improve diffusion classifier performance directly,
highlighting the distinct advantage of our self-improving minority preference optimization framework.

\subsubsection{Effect of KL Regularization.}
Table~\ref{tab:ablation-acc-only} reports the impact of KL regularization on accuracy.
Introducing the KL term yields consistent improvements in both minority and majority groups
(approximately +2\% and +0.6\%, respectively).
This demonstrates that KL regularization plays a crucial stabilizing role during minorit policy optimization by preventing the
policy from drifting too far from the pretrained diffusion prior.
By constraining the update magnitude, the model is able to explore minority regions while preserving
majority representations and overall denoising fidelity.
Without the KL constraint, the policy tends to overfit high-reward states, leading to instability and degraded consistency.

\subsubsection{Effect of Timestep Selection.}
We further examine how the choice of diffusion timesteps affects fine-tuning performance.
As shown in Table~\ref{tab:timestep-ablation}, among the four configurations -- Full (0--50),
Late (30--50), Middle (15--35), and Early (0--20) -- updating only the early
timesteps (0--20) yields the best performance for both minority and majority groups.
This finding is consistent with prior work~\cite{chan2025mgd, xue2025dancegrpo}, which similarly highlights that early denoising steps play a central role in effective policy optimization. Concentrating policy updates at early stages enables effective learning of meaningful semantic features without disrupting structural coherence.
In contrast, applying updates predominantly at later stages leads to weaker gains or even degradation,
likely due to overfitting on local pixel statistics rather than enhancing semantic diversity.

\subsection{Evaluation on Minority Sample Generation}
Although our primary objective is to enhance diffusion classifier performance, 
a natural question remains: \textit{Does the LoRA truly learn minority concepts?} 
To verify this, we attach our Minority LoRA to the diffusion model, generate images using MSCOCO captions, 
and compare both qualitative and quantitative results with the state-of-the-art MinorityPrompt baseline.

\subsubsection{Qualitative Results on Minority Sample Generation.}
As illustrated in Fig.~\ref{fig:qualitative_results}, MiPO enables the diffusion model to generate images that express minority characteristics more clearly and consistently than the DDIM baseline. Across diverse MSCOCO prompts, MiPO produces richer details, more diverse local structures, and visually coherent patterns that reflect minority concepts, demonstrating that the model has genuinely internalized minority-oriented generation rather than relying on prompt engineering.

\subsubsection{Quantitative Results on Minority Sample Generation.}
Table~\ref{tab:main_results} summarizes the quantitative comparison on SD~1.5.  
MiPO improves both perceptual-quality metrics (e.g., CLIPScore, HPSv3) and minority-sensitive metrics (Minority Score, JEPA) over the DDIM baseline, indicating stronger minority representation while maintaining high visual fidelity.  
Importantly, MiPO is fully \textit{prompt-adaptive} and does not require any per-prompt optimization, unlike MinorityPrompt~\cite{um2025minority}, which optimizes a special token for each input prompt.  
Despite this difference, MiPO achieves comparable minority-sensitive scores, demonstrating that our minority policy optimization effectively generalizes across prompts without additional optimization.

\begin{table}[t]
\centering
\caption{\textbf{Human preference evaluation on minority samples.}  
We compare the baseline DDIM sampler, MinorityPrompt, and our MiPO using ImageReward and HPSv2. 
While MinorityPrompt promotes minority-oriented generation, it noticeably lowers human preference scores. 
In contrast, MiPO achieves the best ImageReward score and remains close to the baseline on HPSv2, suggesting a better balance between minority-oriented generation and perceptual quality.}

\resizebox{0.5\linewidth}{!}{
\begin{tabular}{lcc}
\toprule
Method & ImageReward $\uparrow$ & HPSv2 $\uparrow$ \\
\midrule
Baseline                 & \underline{0.184} & \textbf{0.254} \\
MinorityPrompt & -0.026            & 0.241 \\
\rowcolor{gray!10}
Ours (MiPO)              & \textbf{0.228}    & \underline{0.252} \\
\bottomrule
\end{tabular}
}

\label{tab:appendix-imagereward-hpsv2}
\end{table}
\label{Appendix.B.4}

\begin{table}[t]
    \centering
    \caption{\textbf{Zero-shot classification accuracy on LabelME and VOC2007.}
    Due to domain gap (\eg, multi objectes) or inconsistent annotations in these datasets, improvements are less stable 
    compared to clean benchmarks. A detailed analysis is provided in the Appendix.}
    \small
    \begin{tabular}{l|cc}
    \toprule
    Model & LabelME & VOC2007 \\
    \midrule
    SD 1.5         & \textbf{62.90} & \textbf{81.84} \\
    \rowcolor{gray!10}
    Ours           & 55.28 & 76.81 \\
    \midrule
    SD 2.0         & \textbf{63.84} & \textbf{82.38} \\
    \rowcolor{gray!10}
    Ours           & 58.79 & 80.78 \\
    \bottomrule
    \end{tabular}

    \label{tab:sd_version_labelme_voc}
\end{table}
\subsection{Human Preference Evaluation}

We further evaluate human preference using ImageReward~\cite{xu2023imagereward} and HPSv2~\cite{wu2023human}. 
Table~\ref{tab:appendix-imagereward-hpsv2} compares images generated by the baseline DDIM sampler, MinorityPrompt, and  MiPO. 
While minority-oriented generation can improve coverage of low-density regions, it may also introduce a trade-off: aggressively pushing samples toward minority regions can reduce visual fidelity or human preference alignment. 
This trend is reflected in the results, where MinorityPrompt produces minority-biased samples but obtains lower ImageReward and HPSv2 scores.

In contrast, MiPO maintains stronger human preference scores while promoting minority-oriented generation. 
Notably, MiPO achieves the highest ImageReward score, surpassing both the baseline and MinorityPrompt, and remains very close to the baseline on HPSv2. 
These results suggest that MiPO does not simply increase minority characteristics at the cost of perceptual quality; rather, it provides a better balance between minority-oriented generation and human-preferred visual quality.

\subsection{Limitations}
\label{sec:limit}
While MiPO consistently improves zero-shot diffusion classification across most datasets, 
it does not yield gains uniformly. Because our method relies solely on caption-based self-improving policy optimization 
without access to additional image data, certain datasets exhibit weaker or unstable improvements. 
As shown in Table~\ref{tab:sd_version_labelme_voc}, datasets such as LabelME and VOC2007 contain a number 
of multi-object or ambiguous scenes, which introduce uncertainty in zero-shot evaluation. 
A detailed analysis of these cases is provided in the Appendix.

Another limitation arises in minority-oriented image generation. 
Although our policy effectively increases minority representation, it introduces an inherent trade-off between 
minority distinctiveness and overall perceptual fidelity. 
When the optimization places stronger emphasis on minority reward, the model may generate images that 
better reflect minority characteristics but drift slightly from the high-level aesthetic or semantic alignment 
of the pretrained diffusion prior. Conversely, enforcing stronger regularization preserves perceptual quality 
but limits the strength of minority expression. 
Balancing this trade-off remains an open problem and a promising direction for future work.

\section{Conclusion}
In this paper, we propose self-improving diffusion classifiers with Minority Preference Optimization (MiPO), which enhances the performance of diffusion classifiers by self-generating training samples guided by minority-preference rewards. To achieve stable and prompt-adaptive optimization, we fine-tune the diffusion model with LoRA using Group Relative Policy Optimization (GRPO), allowing effective policy updates across diverse prompts. As demonstrated on five standard benchmarks, MiPO successfully improves overall diffusion classifier performance without requiring any external model or image data. In summary, we show that the perception capability of diffusion classifiers is inherently focused on majority groups and demonstrate that enhancing minority group representation can improve the model's overall recognition capability.

\section*{Acknowledgements}
This research was partly supported by the Institute of Information \& Communications Technology Planning \& Evaluation (IITP) grants funded by the Korea government (MSIT) (IITP-2026-RS-2026-25543726, Leading Generative AI Human Resources Development, 23\%; No. IITP-2026-RS-2023-00258649, Information Technology Research Center (ITRC), 17\%; No. RS-2019-II190079, Artificial Intelligence Graduate School Program (Korea University), 1\%), the Culture, Sports and Tourism R\&D Program through the Korea Creative Content Agency (KOCCA) grant funded by the Ministry of Culture, Sports and Tourism in 2024 (Project Name: International Collaborative Research and Global Talent Development for the Development of Copyright Management and Protection Technologies for Generative AI, Project Number: RS-2024-00345025, 10\%), and the National Research Foundation of Korea (NRF) grants funded by the Korea government (MSIT) (No. RS-2025-00562437, 10\%; No. RS-2024-00341514, 20\%; No. RS-2025-02263628, 19\%).

%
%
\bibliographystyle{splncs04}
\bibliography{main}
\end{document}